# Semantic rule Web-based Diagnosis and Treatment of Vector-Borne Diseases using SWRL rules

Ritesh Chandra, Sadhana Tiwari, Sonali Agarwal, Navjot Singh

rsi2022001@iiita.ac.in, rsi2018507@iiita.ac.in, sonali@iiita.ac.in, navjot@iiita.ac.in

Indian Institute of Information Technology Allahabad, Prayagraj, India

**Abstract**

Vector-borne diseases (VBDs) are a kind of infection caused through the transmission of vectors generated by the bites of infected parasites, bacteria, and viruses, such as ticks, mosquitoes, triatomine bugs, blackflies, and sandflies. If these diseases are not properly treated within a reasonable time frame, the mortality rate may rise. In this work, we propose a set of ontologies that will help in the diagnosis and treatment of vector-borne diseases. For developing VBD's ontology, electronic health records taken from the Indian Health Records website, text data generated from Indian government medical mobile applications, and doctors' prescribed handwritten notes of patients are used as input. This data is then converted into correct text using Optical Character Recognition (OCR) and a spelling checker after pre-processing. Natural Language Processing (NLP) is applied for entity extraction from text data for making Resource Description Framework (RDF) medical data with the help of the Patient Clinical Data (PCD) ontology. Afterwards, Basic Formal Ontology (BFO), National Vector Borne Disease Control Program (NVBDCP) guidelines, and RDF medical data are used to develop ontologies for VBDs, and Semantic Web Rule Language (SWRL) rules are applied for diagnosis and treatment. The developed ontology helps in the construction of decision support systems (DSS) for the NVBDCP to control these diseases.

## 1. Introduction

VBDs are a kind of health issue caused by pathogens spread by arthropods such as triatomine bugs, mosquitoes, blackflies, sand flies, tsetse flies, ticks, and lice [1]. Vectors are biological organisms that can spread infectious disease from one person to another or from one animal to another. VBDs account for around 17% of all illnesses among all infectious diseases. As per one of the reports of the World Health Organization (WHO), more than 1 billion illnesses and over 1 million fatalities happen per year due to VBDs [2]. In India, VBDs are being controlled and prevented through the NVBDCP, which was introduced in 2003-04 by the Government of India. The NVBDCP [3] is formed by combining the National Filaria Control Programme, the Kala-azar Control Programme, and the National Anti-Malaria Control Programme. It also includes dengue and Japanese B encephalitis. NVBDCP receives funding from the World Bank and the Global Fund ATM (GFATM) to focus its efforts primarily in most endemic areas to control malaria and eliminate kala-azar, which has a negative impact on poor people who live in dense, untidy, and unsanitary housing [4]. WHO is also a significant partner that provides necessary support, rules, guidelines, and technical advice to the programme. In the present scenario, all countries over the world are facing a shortage of doctors, especially India. The majority of people are suffering greatly as a result of their lack of knowledge about proper medical treatment and checkups. The proposed model is useful to cope with this problem. We deployed this model in the form of an app and website in remote areas, which reduces the patients' dependencies upon the doctors and helps the people avoid paying huge amounts to the doctors unnecessarily.

A Decision Support Systems (DSS) [5] is a computer-based information system that integrates models and data to handle unstructured or semi-structured problems with multiple user engagements via a friendly user interface. DSS plays an important role for both the doctor and the patient. It not only assists physicians in diagnosis and treatment but also improves healthcare remotely, affecting the quality of life of patients. The semantic web is an effective option for knowledge sharing and representation in order to improve one's expertise. One of the pillars of the semantic web is ontology. It is defined as "a technology for knowledge representation that has been adopted". It functions as a domain-specific dictionary, defining objects, properties, and the relationships between them.

Ontology is basically a data model where knowledge can be represented by using concepts related to any domain and defining relationships among the concepts. Nowadays, ontologies are utilized in the field of information science to accomplish a variety of activities, such as improving user-machine communication. It also makes use of any pre-existing data model or knowledge schema. As a result, the fundamental concept of the semantic web has grown to a higher level, and several types of ontologies have been designed. Among the several categories and classifications, Basic Formal Ontology (BFO) [6] is the only one that supports reasoning in addition to the general Open World Semantics, which is followed by all ontologies.

This work deals with the text medical data that is collected by doctors' handwritten notes, mobile medical applications, and websites. Then all this text data is combined, and meaningful text is extracted from it using some text-based algorithms like OCR, spell checker, and NLP [7]. Then convert this text data into an RDF medical data [8] with the help of NLP [9] and the PCD ontology [10]. Then develop a formal ontology using this RDF medical data, the NVBDCP guidelines, and SWRL rules [11] for the diagnosis and treatment of VBDs. This work's ultimate purpose is to digitize the NVBDCP by combining the ideas of the DSS [5] and the Web Ontology Language (OWL) [12]. The following points illuminate the major findings of this work:

- The text-based medical data is transformed into Resource Description Framework triples to increase its quality and reuse in the future.
- To develop a rule-based diagnosis and treatment system for VBD's patients, a set of rules can be defined with the help of classes, properties, and persons using Semantic Web Rule Language.
- The developed Knowledge Driven DSS aids new meaning in actual decision making with the support of facts, rules, and procedures.
- Design BFO ontology for better understanding of VBDs guideline, precautionary measure and working process of NVBDCP.
- Form a new text extraction model based on natural language processing for retrieving meaningful medical text data.

The remaining part of this work is arranged as follows: Section 2 provides a glimpse of existing related work about knowledge representation through ontology. Section 3 explains the proposed methodology of construction of VBD's ontology. Section 4 discusses ontology development based on BFO using NVBDCP guidelines, semantic web rule language (SWRL), usage of SWRL rules in diagnostic classification ontologies, and a practical framework view of diagnosis and treatment of VBDs. Section 5 presents the results of the metric-based evaluation of ontology, and Section 6 reports the conclusion and future scope.

## 2. Related Work

Many studies have been previously done in the case of vector borne disease identification, precaution, and treatment. These studies highlighted the potential research gaps and interest in the diagnosis of diseases caused through the transmission of vectors. Semantic web-based disease modeling is a relatively new term that has piqued the interest of researchers and medical practitioners dealing with diseases spread by mosquitoes, fleas, and other bacteria.

Topalis, P. et al. [13] developed a tool that is extremely beneficial for the malaria community due to its efforts to reduce the worldwide malaria burden effectively. They propose Infectious Disease Ontology—Malaria (IDOMAL), the first operational malaria ontology whose objective is to design a standard language for the community that computers and dedicated software can both understand. The IDOMAL [14][15] contains almost two thousand terms. This ontology captures multiple aspects of disease, such as clinical, epidemiological, and vector biology. Some other works provide the experimental evaluation of the disease control system, in which eight different diseases are considered, including dengue, malaria, cholera, diarrhea, influenza, meningitis, leishmaniasis, and kala-azar. The disease may be detected on the basis of primary symptoms, and their relationships will be valuable for developing a biomedical knowledge base (e.g., a disease ontology) for e-health and disease surveillance systems [15][16][17].

The most common diseases afflicting Indonesian society are classified as tropical diseases, such as malaria, leprosy, and lymphatic filariasis. Using the Resource Description Framework (RDF) serialization form, Semantic Web can display data relationships in Bahasa Indonesia [18]. Table 1 provides a summarized list of existing researches performed on various infectious diseases using ontology.

Table 1: Existing research performed on various infectious diseases using ontology

| Targeted disease | Objective | Technology used | Result |
| --- | --- | --- | --- |
| Based on infectious diseases like bacteria, fungus etc. [19] | To construct a disease intelligence system, acquire and synthesize fragmented illness knowledge obtained from diverse sources. | Ontology creation and analysis. | Class hierarchy with illness ontology features. |
| Based on dengue[20] | To integrate and navigate many types of information in the context of dengue sickness. | Ontology creation and text mining. | Dengue data integration and sharing generic infrastructure. Explanation of dengue serotype trends. |
| Based on dengue[21] | To provide a system for detecting similarities among dengue-infected people in order to effectively manage the outbreak. | Creation of ontology-dependent domain thesaurus and case-based reasoning for similarity identification. | In terms of accuracy and error rate, the framework surpasses the User-based Pearson Correlation Coefficient (UPCC) and Item-based Pearson Correlation Coefficient (IPCC) methods. |
| Based on dengue[22] | Imputing missing data to improve the predictive ability of data mining techniques. | Semantic data imputation based on ontology inference for DF epidemic data. | Experiment results show that semantic data imputation outperforms statistical techniques. |
| Based on Break bone fever[23] | To broaden the knowledge base preventing and controlling vector-borne disease. | IDODEN ontology creation. | DF taxonomy including biological, epidemiological, and clinical characteristics. |

After reviewing numerous papers in the literature, it is evident that no one has discussed combining complete VBDs into a single platform. Only a few works show knowledge representation through ontologies such as the IDODEN and IDOMAL ontologies. In the proposed work, a new ontology is constructed for VBDs that encompasses all vector-borne illnesses under one roof to make it more efficient. This ontology is operational to support the diagnosis and therapy of VBDs with the help of SWRL rules. The text data collected from patients' databases, doctor's handwritten notes, and test reports of patients will be extracted using OCR and corrected with the help of a spell checker. This work also helps in making an RDF medical data using the PCD ontology, which is based on text entity extraction through NLP.

## 3. Proposed Methodology

### 3.1 Dataset Description

The electronic health records taken from the Indian Health Records website are used for making ontology [24], which includes details of patients' age, smoking habits, drug use, previous diseases, and so on. This data is maintained by CDAC Mohali. The text data used in this work is taken from the Vector Borne Disease Control and Surveillance mobile application [25] and doctors' handwritten notes, in the form of images and documents (i.e., pdf, doc, etc.). Many other health guidelines have been formed with the help of the NVBDCP guidelines [3].

### 3.2 Structure of the proposed model

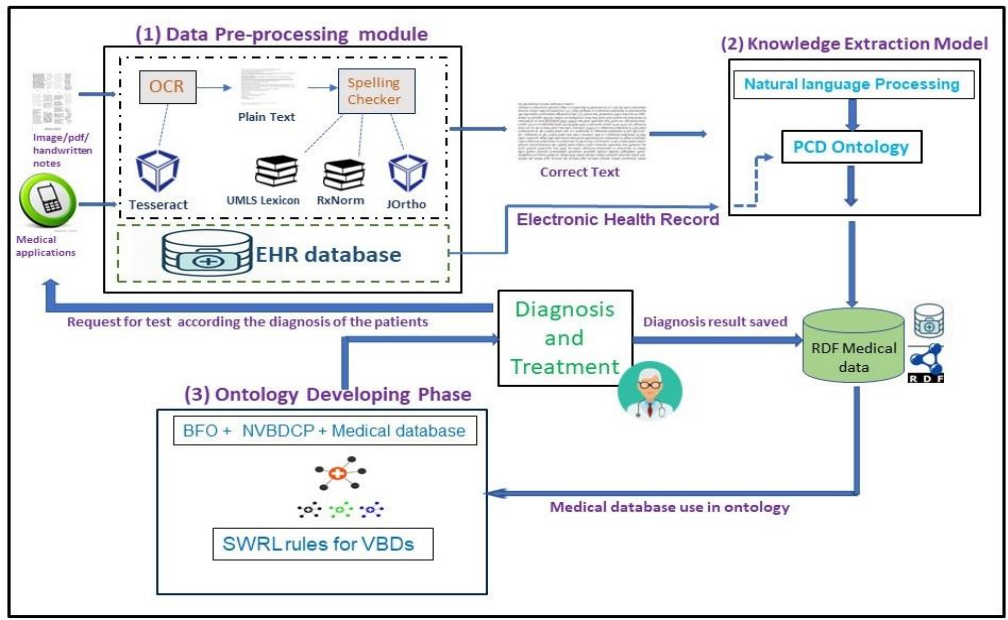

Figure 1: Complete architecture of the proposed model

### 3.2.1 Module for data pre-processing

The doctor's written clinical notes and medical mobile application data [25] are fed into the data pre-processing module so that they can be processed in the correct text data format. Two processes are used to convert this data into correct text data, which is then given to NLP for knowledge extraction from the text data: optical character recognition [26] and spell checking [27]. Electronic health records are given to the PCD ontology after preprocessing.

### a) Optical Character Recognition

The system through which scanned images of any document, such as handwritten or typed text, can be converted into machine-readable and editable text documents is defined as OCR. Handwriting recognition refers to a computer's capacity to recognise and comprehend handwritten input from a variety of sources, including paper files, touch screens

[28], and other kinds of devices. Typed or handwritten documents are available in a wide range of typefaces and styles. The three sorts of writing styles are continuous (cursive) text, distinct (handprint or boxed) text, and mixed text. In the ontology-based clinical information extraction (OB-CIE) process [8], a pen is used by the physician to record the details about a patient's visit on a paper note; after this, the paper is scanned and stored on the computer in image format. Handwritten text is recognized by the OCR component, which converts it into a text file that may be changed. The OCR component is built using the Tesseract OCR engine [29], which is a new open-source OCR engine developed by HP, and it is rewarded as the most powerful and accurate OCR engine among the existing ones. The Tesseract OCR engine is integrated with OB-CIE using Tess4J [8]. Tess4J works as a Java JNA wrapper for the Tesseract OCR API, which is licensed under the Apache License, version 2.0.

**b) Spelling Checker**

Output text generated by the OCR system can produce typographical errors against the recognized text if different writing styles are present in the scanned copy of a handwritten file. Before diving deeper into the next step, it must be ensured through spell checking that each concept is correct. The two steps followed in the OB-CIE spell checker are: firstly, it figures out the misspelled concepts and translates these concepts into correct form. This spell checker unit is built using JOrtho (Java Orthography) [30]. The dictionary of JOrtho (a Java-based open-source library) is established with the help of the Wiktionary project, which has 5,836,006 entries over 3800 languages, including English definitions [8]. The spell-checker unit of the system was developed using the JOrtho dictionary. The spell checker verifies from the dictionary whether each term is correct or not. If any particular term isn't discovered, then it is considered a misspelled term by the spell checker, and it is being highlighted. Secondly, the checker consults the dictionary to provide a list of relevant term options, which are then ranked to allow the user to select the most appropriate. If the term is correct but not in the dictionary, it could be added by the user.

**3.2.2 Knowledge extraction based on Natural language processing**

The ability of a machine to collect data can be improved by detection based on natural language comprehension. In particular, the semantic meaning of phrases is taken into account as a factor to aid the machine in intent identification. Because of its scalability, a comparison of semantics between input and prepared sentences is seen as a good option for detecting human intentions. The cosine distance between sentences can be used to calculate a value of semantic similarity based on their distribution of representation. A probabilistic-based neural-net language model to assist machines in learning helps in the distributed representation of words, which is a well-known idea in natural language processing.

**a)      Sentence boundary detection**

To determine the ending of a sentence, the OpenNLP Sentence Detector examines the punctuation character. The beginning of a sentence is presumed by the first non-whitespace character, and the end of the sentence is recognized using the last non-whitespace character. When the boundaries of a sentence have been detected, each subsequent sentence will be written on a single line.

**b)   Tokenization**

This is the process of applying various steps to the token, like lemmatization, stemming, conversion of uppercase text to lowercase, stop word removal, and efficiently finding the portion of text [31] [32]. Each sentence is segmented into tokens using OpenNLP tokenizers, which are basically a string of whitespace characters [33].

### c) Stop words removing

To improve the identification of concepts and their relationships during the information extraction phase, stop words must be removed from the text created by the tokenization process. A collection of common stop words that appear frequently in physicians' notes was painstakingly compiled.

### d) Part of speech tagging (POS)

In the NLP module, POS tagging is used to detect the tag of every word, and some sets of rules are used to eliminate garbage verbs from phrases like "started" and "sought." In the information extraction module, removing these verbs will allow you to focus on detecting noun terms. The tokens and their correlative word types can be described using OpenNLP POS Tagger, and the right POS tag can be predicted using a probability model [33].

### e) Sentence parsing

Noun phrasing is a well-known NLP technique that is used to determine whether the standard keywords can be combined to enhance the targeted information's quality [34]. A parse tree is created across each textual input during the parsing step, which represents a hierarchical structure that describes the grammatical structure of a sentence. The OpenNLP parser is used to find noun phrases in any text by extracting combinations of words that are tagged with NP. Fever, cough, toxemia, BMI, and other medical terms were generally expressed in the form of phrases or words like acute sinusitis, nocturnal enuresis, loss of appetite, etc. [35]. Figure 2 depicts a complete illustration of the working of NLP module.

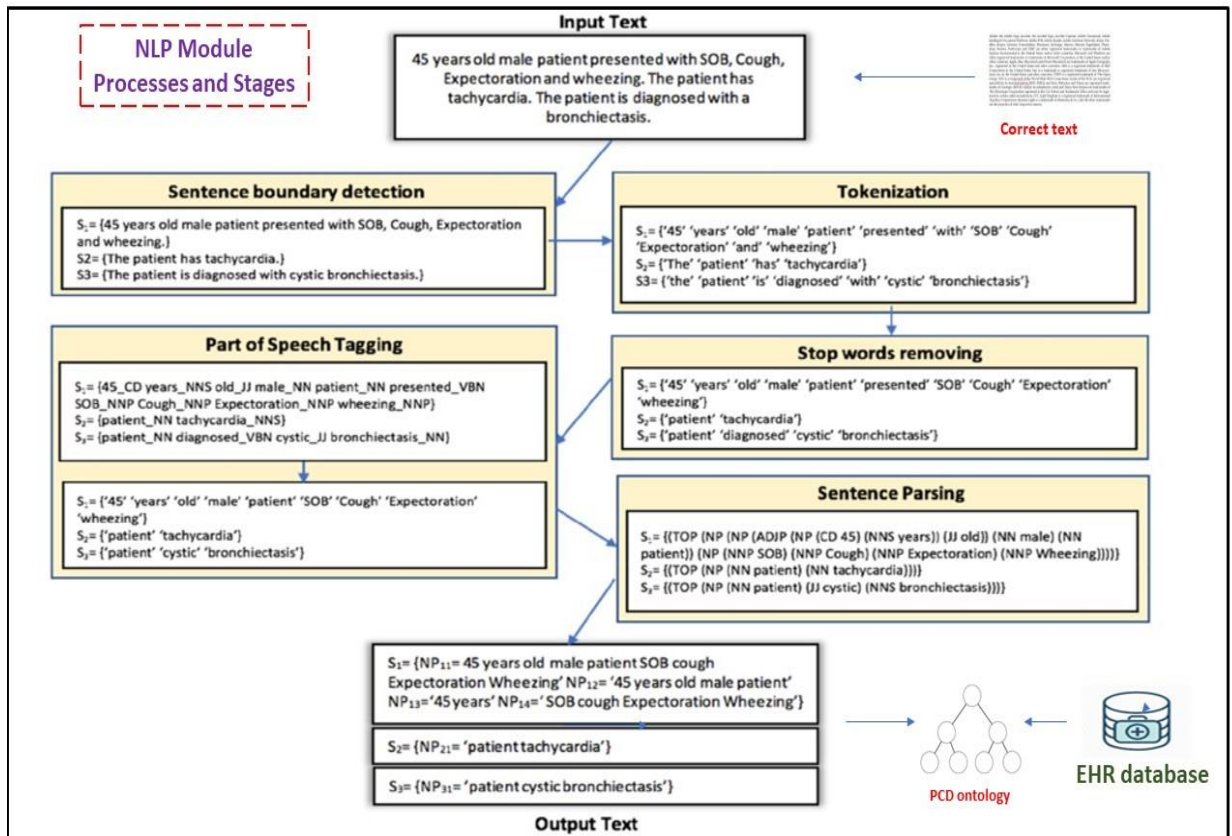

Figure 2. The NLP working process and its stages

### 3.2.3 PCD Ontology

Patient Clinical Data (PCD) is an ontology that describes the components of EHR clinical data. It represents clinical principles relating to healthcare activities that occur during a patient's visit. The output of NLP Correct text will pass intently by extracting the appropriate information based on the domain we choose. For extracting meaningful information, many approaches are available, like Word2Vec, FastText [36], etc. Then map this exact word through the PCD ontology [37] and convert this word into a Resource Description Framework (RDF) database [38] frame view as shown in figure 3. Further processes are handled by VBD's ontology, which is shown in Figure 1.

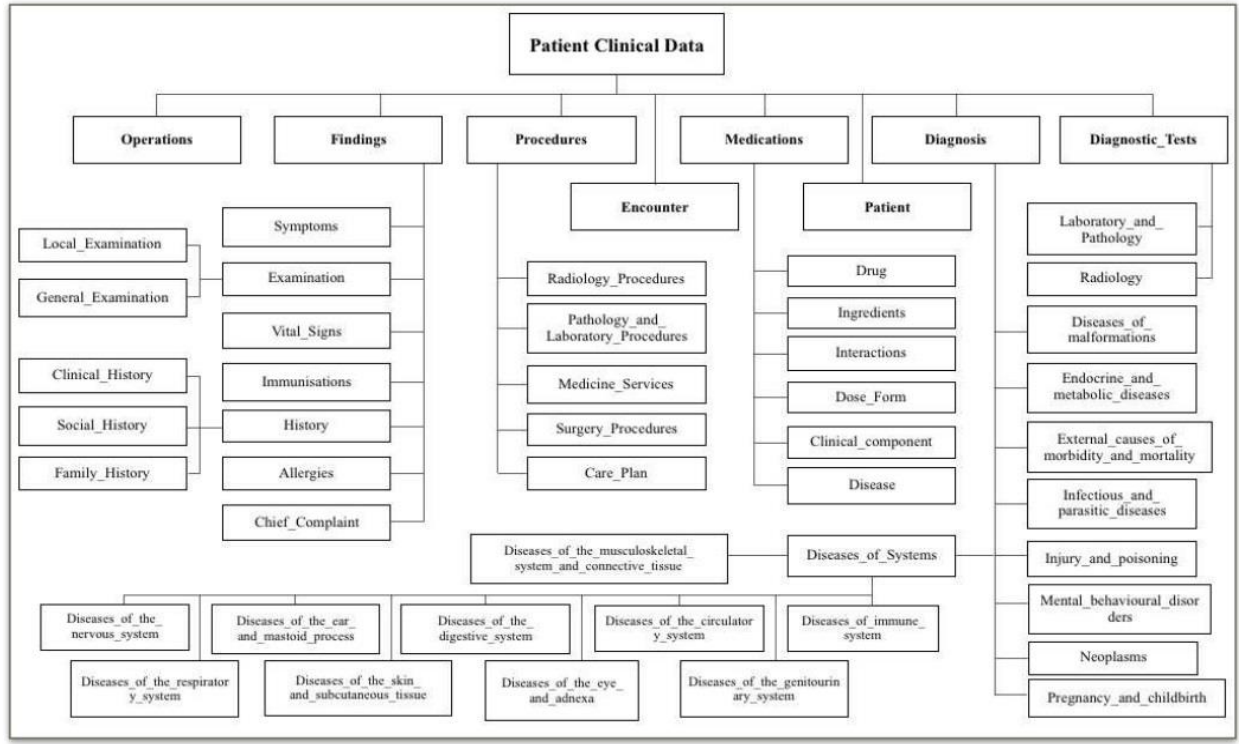

Figure 3. The PCD ontology's top-level class hierarchy contains patient clinical information. [39]

## 4. Ontology Development

Doctors must ask patients for more symptoms in order to aid accurate information in the diagnosing process during medical check-ups in order to create a basic formal ontology for VBDs. This technique necessitates a doctor's medical knowledge in order to recognize the signs that must be discovered. Recognition of disease by following symptoms is very tricky using rules in software programmes; nevertheless, this still has drawbacks because every disease has too many symptoms, and some of them may overlap, which imposes a challenge in system development. This barrier could be overcome by using ontology, a semantic database that can represent the relationship between symptoms and diseases. An ontology-based strategy is proposed for automatically detecting the necessary symptoms during the conversation of patients by following the procedure of a medical checkup. The system can detect symptoms in conversations and forecast the required subsequent symptoms by incorporating a neural network into the ontology database.

## 4.1 Basic Formal Ontology Based Representation of Vector Borne Diseases

This section focuses on representing NVBDCP using an upper-level ontology such as Basic Formal Ontology (BFO) [40]. BFO is designed to be domain-neutral in order to facilitate the interoperation of what are known as "domain ontologies" developed on its foundation and therefore to support uniform data annotation across multiple domains. BFO is a method of describing a basic entity that does not particularly focus on a problem area; it is commonly used in the representation of biomedical data. In this work, the domain of interest is vector-borne disease, and we have used BFO to represent NVBDCP (as shown in Figure 4).

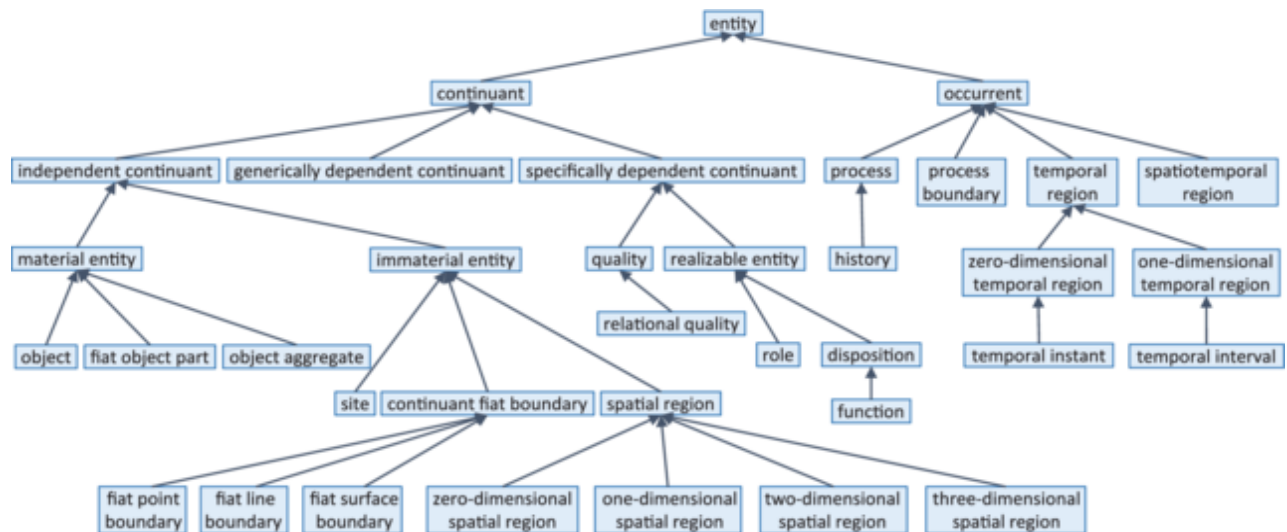

Figure 4. BFO[1] structure for VBDs

BFO divides any entity into two types: continuant entities and occurrent entities. A continuant is a thing that remains the same or doesn't change through time. The term "current" entity refers to something that varies through time. The "continuant" entity is further divided into two subcategories, namely independent continuant, and dependent continuant. Independent continuation can be either generically or specifically dependent continuation. There are four subclasses of occurrent entities: process, process boundary, temporal region, and spatiotemporal region. In this work, many concepts (entities) have been collected that are related to our domain of interest, i.e., VBDs and NVBDCP programmes. Then we divided these ideas into nouns and verbs. Nouns form the basis for classes, and verbs form the basis for object properties and occurrences. Then we have put those concepts in the BFO structure at their appropriate positions. Then we have defined data properties and object properties for those concepts, and after that we have made rules for the diagnosis and treatment of VBDs by relating those concepts using SWRL.

### 4.1.1 The implementation of NVBDCP ideas as a Continuant entity

The organizational and operational structure of the NVBDCP is separated into several tiers, like state level, regional level, and national level. The state level is further divided into district level, sub-district level, and many more levels. Every level comprises actors responsible for implementing NVBDCP. The Directorate of NVBDCP, Ministry of Health and Family Welfare (MoHFW, Govt. of India), Directorate General Health Services, additional directors, joint directors, research officers, and other staff members such as accountants, data entry operators, and others work at the national level. At the regional level, the regional director, entomologists, and other entomology staff are associated; at the state level, the state programme officer (SPO) (for VBDs), the deputy director, entomologists, secretaries, and other staff are involved; and at the district level, the District VBD Control Officer (DVBDCO), the malaria inspector, the Malaria Technical Supervisor (MTS), the Kala-azar Technical Supervisor (KTS), and other support staff are working. Apart from that, at subdistrict and below-level MO-PHC, other health staff like Accredited Social Health Activist (ASHA), Multi-Purpose Health Worker (MPHW), etc. are connected. All these posts are occupied by a specific person and hence treated as "roles." "Role" comes under a realizable entity because it can be realized (as visualized in figure 5 (1)).

---

[1] https://www.iso.org/obp/ui/#iso:std:iso-iec:21838:-2:ed-1:v1:en

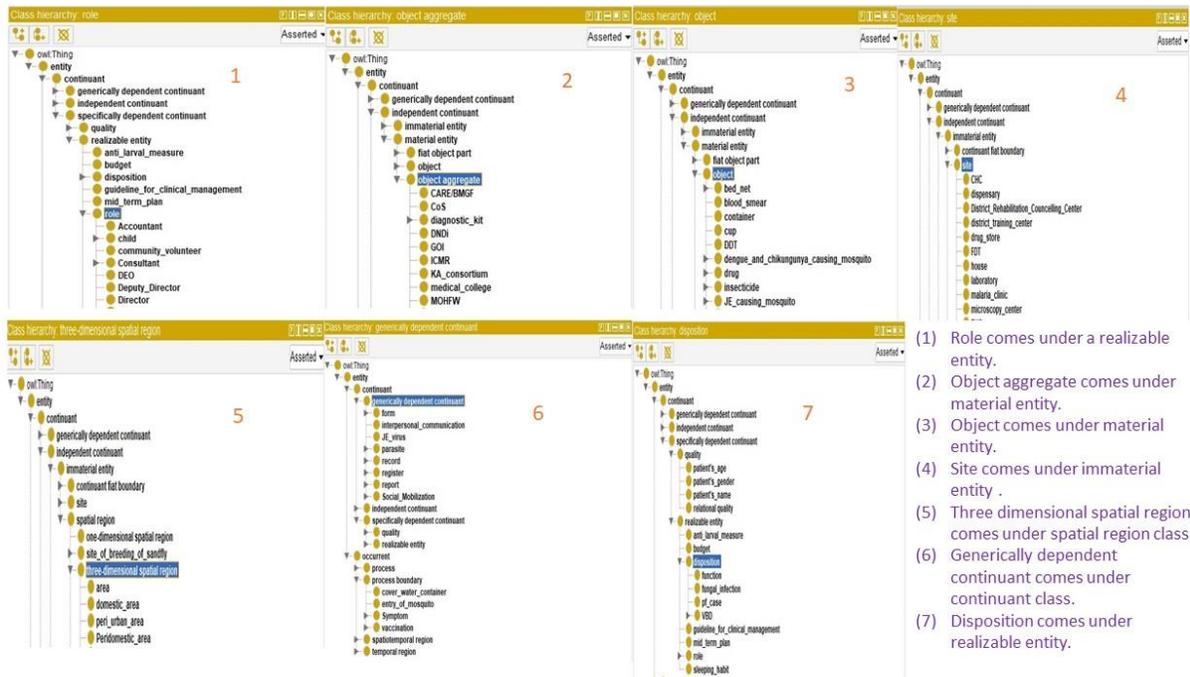

Figure 5. Classes of role, object aggregate, objects, site, three-dimensional capital region generically dependent and disposition

When it comes to combating vector-borne diseases in India, the NVBDCP works in tandem with the Ministry of Health and Family Welfare (MoHFW), an autonomous government agency. It employs a large number of people and organizations. As a result, it's possible to implement it as an **"object aggregate" (figure 5.(2))**. Because an object aggregate is a material entity, it will retain its identity even if some of its components are added or removed. In medical schools and hospitals, for example, if any staff or departments are added or removed, the institution retains its identity. So, it can be considered an object aggregate. Any organization, such as NGOs (non-governmental organizations) and research institutes such as the Indian Council of Medical Research (CMR) [41], the National Institute of Malaria Research (NIMR) [42], and others, can be represented as an object aggregate. Considering the interest of our domain, i.e., VBDs, a patient has its own importance in our domain. Objects are those that have some special importance in the relevant area.

Various VBDs such as malaria, dengue, filaria, chikungunya, and JE covered under the NVBDCP are spreading through mosquitoes, while the Kala-Azar VBD is transmitting through sand flies. So, the patients, mosquitoes, and sand flies can be considered "objects," which is shown in figure 5(3). Bed nets (ITNLLINs) for mosquito protection, DDT, medications, insecticides, diagnostic kits, blood smears (needed for detecting cases), and other items are also important in this study; these items are considered "objects." Independent continuants can be categorized as immaterial entities or material entities. A continuous fiat boundary, site, and spatial region comprise an immaterial entity. The term "site" refers to a three-dimensional immaterial entity that is bounded by a physical entity. In other words, a "site" is reliant on material elements. PHC (Primary Health Center), CHC (Community Health Center), regional training centers, malaria clinics, and district training centers rely on the Family Welfare and Ministry of Health, Government of India, for funding, providing guidelines, and all other relevant resources. Hence all these can be recognized as **"sites" (as in** figure 5(4)). Laboratories, microscopy centers, and drug stores are materially dependent on hospitals and medical colleges.

A **'spatial region'** is an immaterial thing that is defined with regard to some reference frame and is a continuous portion of any space. For example, domestic areas, peri-urban areas, peri-domestic areas, rural regions, and residential blocks can be described in terms of a reference frame such as district, state, or nation. These all can be recognized as a **'three-dimensional spatial region'** under the spatial region as depicted in figure 5(5). Various forms used in hospital or medical systems such as patient transfer form, Laboratory form, malaria case investigation form etc. requires several data like name, gender, age etc.

It depends on one or more than one concept. This situation can be considered as '**generically dependent continuant**' (as mentioned in figure 5(6)). Similarly, registers (spray registers, stock registers, etc.), records (laboratory record), reports (laboratory test report, district annual planning report, etc.) and results (laboratory test result) are all considered 'generically dependent continuants' under the continuant entity concept. Because it does not take any additional processing to realize, for instance, anyone can identify a patient's gender simply by looking at him, personal facts like the patient's name, gender, and age can be identified as **"quality"**. Different VBDs viz. Malaria, Dengue, Chikungunya, Filaria, JE and kala-azar are realized as '**disposition**', because it cannot be identified by seeing the physical appearance of a patient. Disease can be described as the state of an organism which as a result shows one or more biological system problems.

**4.2 SWRL rules for diagnosis and treatment of VBDs under NVBDCP guidelines**

SWRL (Semantic Web Rule Language) [43] is used to define rules and logic for semantic web. In the proposed work, we have defined several concepts related to NVBDCP and by following NVBDCP guidelines we have established relationships among those concepts using SWRL, to define rules for diagnosis and treatment of different vector borne diseases. These rules are the core part of our work which will be helping in decision making to identify symptoms of particular disease and take suitable action for diagnosis and treatments.

Patients are considered as an object, as a DSS is developed in this work for vector-borne disease, and here VBDs are realized as dispositions because they can be realized, and they can change the physical appearance of a patient. The fact that disease can exist without any proper manifestations (i.e., without realization of the disposition) and that it can appear in a variety of ways is explained by considering sickness to be a disposition (dependent, for example, on the presence or absence of symptom-suppressant drugs). VBDs have their own diagnosis and treatment plan defined by the NVBDCP. The suggested system makes heavy use of rules to direct the handler to perform the appropriate actions based on the patient's situation. The knowledge base that has been built is unable to infer new information. To extract the relevant knowledge, some rules must apply to the knowledge base. The developed system focuses on patient care and management for the various VBDs, such as:

      a) .Lymphatic Filariasis
      b). Chikungunya
      c). Dengue
      d). Malaria
      e). Kala-azar
      f). Japanese Encephalitis (JE)

The diagnosis process used by NVBDCP utilizes Semantic Web Rule Language (SWRL), which is built on Web Ontology Language Description Logic (OWL-DL) [44] and Horn logic, to diagnose and treat patients in the aforementioned groups. The rules were created in SWRL at first and then implemented with the pellet reasoner. SWRL rules given in Table 2 display the NVBDCP criteria for detecting illnesses in people with suspected symptoms.

Table 2: NVBDCP criteria for detecting illnesses in people with suspected symptoms

| S.No. | VBD detection using SWRL rules |
|---|---|
| 1 | patient(?p) ^ has_Fever_WithChills(?p, true) ^ has_Headache(?p, true) ^has_Nausea(?p, true) -> has_SymptomOf_Malaria(?P, true) |
| 2 | patient(?p) ^ has_Fever(?p, true) ^ has_Headache(?p, true) ^has_JointPains(?p, true) ^has_Muscle_Pain(?p,true)^has_Vomiting(?p,true)^has_Hemorrhagic_Manifestations(?p,true)->has_SymptomOf_Dengue(?p, true) |
| 3 | patient(?p) ^ has_Fever(?p, true) ^ has_Headache(?p, true) ^has_MildInfection(?p, true) ^ has_Neck_Stiffness(?p, true) ->has_SymptomOf_JE(?p, true) |
| 4 | patient(?p)^has_Elephantiasis(?p,true)^has_Hydrocele(?p,true)^has_Lymphoedema(?p, true) -> has_Symptom_Of_Filaria(?p, true) |
| 5 | patient(?p) ^ has_Chills(?p, true) ^ has_Fever(?p, true) ^has_Headache(?p, true) ^ has_Joint_Pains(?p, true) ^ has_Rash(?p, true) ^ has_Vomiting(?p, true) -> has_SymptomOf_Chikungunya(?p, true) |
| 6 | patient(?p) ^ has_Anaemia(?p,true)^has_Dry_Skin(?p,true)^has_Recurrent_Fever(?p, true)^has_Weakness(?p,true)^has_Weight_Loss(?p,true)->has_Symptom_Of_Kalaazar(?p, true) |

#### 4.2.1 Development of SWRL rules for identification of different VBDs

As it is well known, chikungunya fever symptoms are quite similar to dengue fever symptoms. Dengue fever does not have hemorrhagic symptoms, and the Chikungunya virus does not cause infectious shock. Malaria is diagnosed using a rapid diagnostic test and microscopic analysis of blood samples (RDT). In villages where microscopic inspection is not possible within one day, RDT is provided by health agencies and health workers such as ASHAs. As a result, treatment can be delivered based on the diagnosis. Tables 2 and 3 list the SWRL criteria for diagnosis and drug selection for malaria treatment, which are based on the NVBDCP diagnosis and treatment model (i.e., diagnosis and treatment for malaria 2013) [45]. Figure 6 represents the procedure of microscopic diagnosis for Malaria disease.

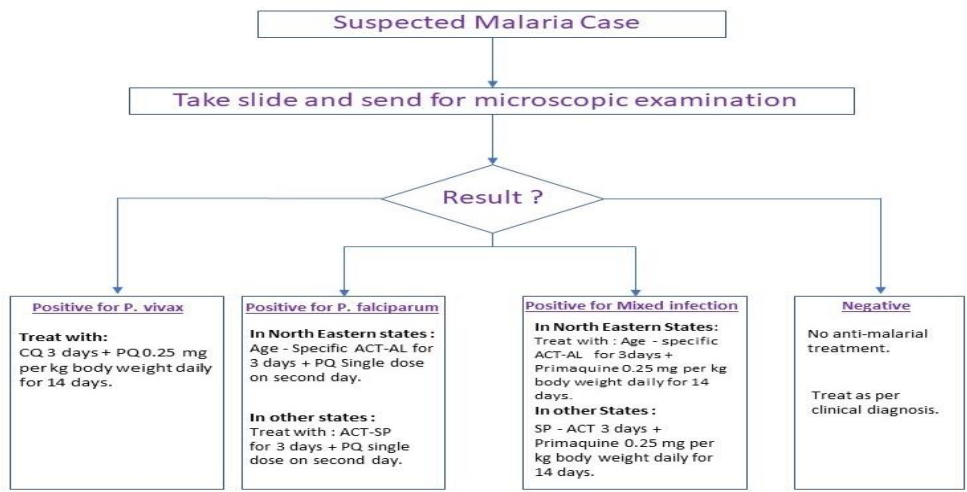

Figure 6: Flowchart of microscopic examination for Malaria [45]

Table 3: SWRL rules for guidelines for diagnosis and treatment of Malaria

| S.No. | When a microscopy result is received within one day, SWRL guidelines applied for diagnosis and treatment of Malaria |
|---|---|
| 1 | Microscopic_Examination(?ME)^patient(?p)^ has_SymptomOf_Malaria(?p, true) -> undergoes(?p, ?ME) |
| 2 | patient(?p) ^ has_ME_Result(?p, ?v1) ^ is_Positive_For_PVivax(?p, true)^swrlb:equal(?v1, "positive") -> has_PVivax_Malaria(?p, true) |
| 3 | patient(?p) ^ has_ME_Result(?p, ?v1) ^ is_Positive_For_PFalciparum(?p,true) ^ swrlb:equal(?v1, "positive") -> has_Falciparum_Malaria(?p, true) |
| 4 | patient(?p) ^ has_ME_Result(?p, ?v1) ^is_Positive_For_Mixed_Infection(?p, true) ^ swrlb:equal(?v1, "positive") ->has_Mixed_Infection(?p, true) |
| 5 | clinical_diagnosis(?cd) ^ patient(?p) ^ has_ME_Result(?lp, ?v1) ^swrlb:equal(?v1, "negative") -> undergoes(?p, ?cd) ^has_Required_Malaria_Treatment(?p, false) |

The treatment process of Malaria disease is shown in figure 7.

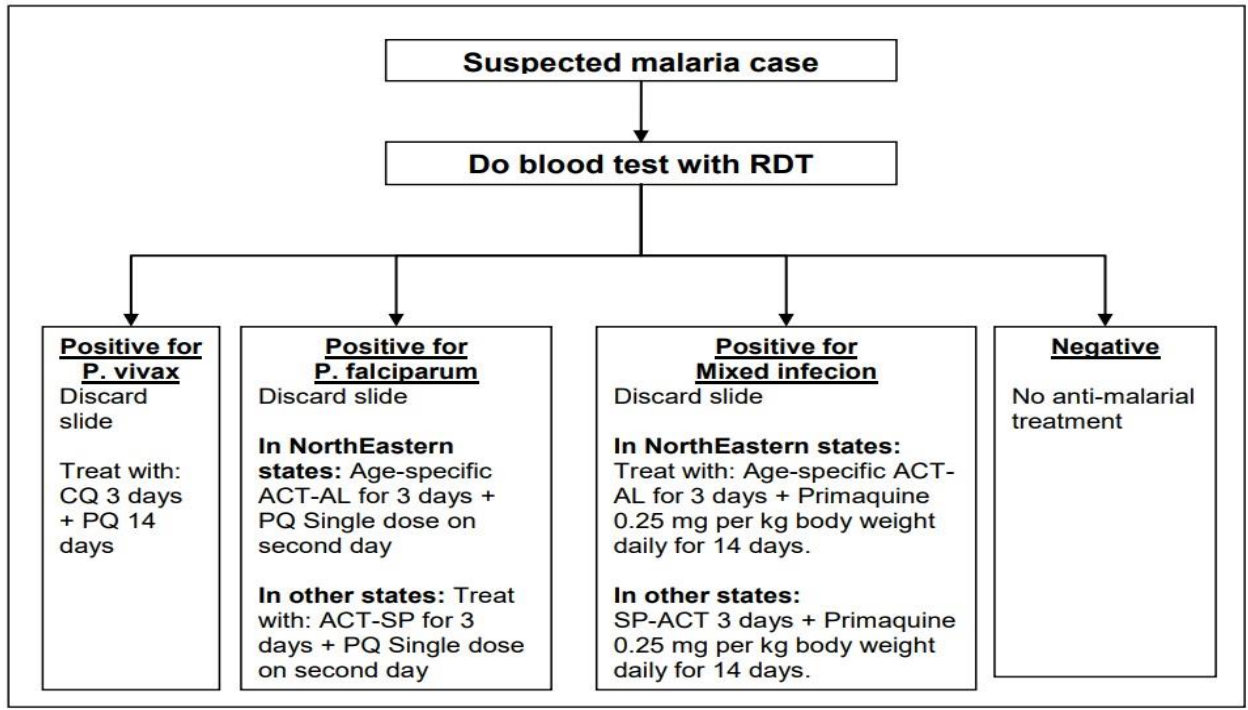

Figure 7: Treatment model for malaria using monovalent RDT [45]

SWRL rules given in table 4, provides guidelines for diagnosis and treatment of malaria with a monovalent RDT.

Table 4: SWRL rules for diagnosis and treatment of malaria with a monovalent RDT

| S.No | When microscopy results are not available within one day and a monovalent RDT is utilized, the SWRL guidelines applied for diagnosis and treatment of malaria |
|---|---|
| 1 | Monovalent_RDT(?v1) ^ rural_area(?ra) ^is_ME_Result_Available_Within_One_Day(?ra, false) -> use(?ra, ?v1) |
| 2 | RDT(?rdt) ^ patient(?p) ^ has_Symptom_Of_Malaria(?p, true) ^ is_Prescribed_RDT(?p, true) -> undergoes(?p, ?rdt) ^prepare_Slide(?p, true) |
| 4 | patient(?p) ^ has_RDT_Result(?p, "positive")^is_Positive_For_PFalciparum(?p, true) -> has_Falciparum_Malaria(?p,true) |
| 5 | ACT-AL(?al) ^ Primaquine(?PQ) ^ patient(?p) ^ belongs_To_North_East_State(?p, true)^ has_Falciparum_Malaria(?p, true) -> is_Prescribed(?p, ?PQ) ^ is_Prescribed(?p, ?al) ^is_Prescribed _For_Duration(?PQ, 1) ^ is_Prescribed_For_Duration(?al, 3)^ is_Prescribed_OnDay(?PQ, 2) |
| 6 | ACT-SP(?sp) ^ Primaquine(?PQ) ^ patient(?p) ^ belongs_To_Other_State(?p, true) ^ has_Falciparum_Malaria(?p, true)->is_Prescribed(?p,?PQ)^is_Prescribed(?p,?sp)^is_Prescribed_For_ Duration (?PQ,1)^is_Prescribed_For_Duration(?sp,3)^is_Prescribed_OnDay(?PQ, 2) |
| 7 | Chloroquine(?cq) ^ patient(?p) ^ has_High_Suspicion_Of_Malaria(?p, true) ^ has_RDT_Result(?p, "Negative")^has_Slide_Result(?p, false)-> isPrescribed(?p, ?cq) ^ is_Prescribed_For_Duration(?cq, 3) |

SWRL rules for screening and patient care suspected with Lymphatic Filariasis symptoms as per the NVBDCP guidelines are mentioned in Figure 8.

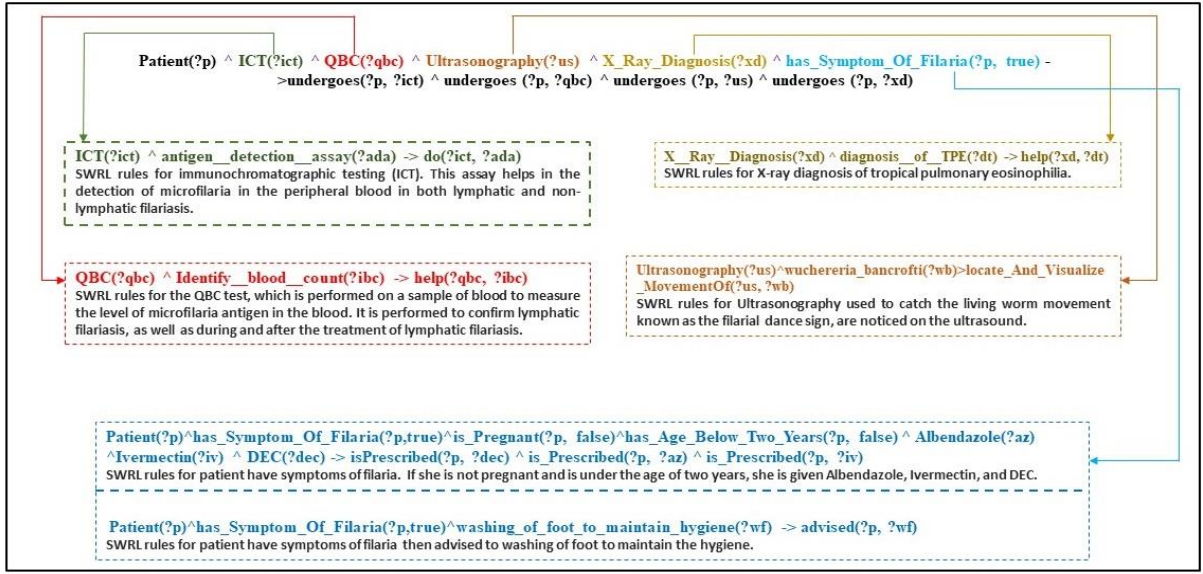

Figure 8: SWRL rules for Filaria with explanation

For the treatment of Japanese encephalitis (JE), there is no particular course of action. Early case management is critical for reducing the risk of complications and mortality. If signs of JE are discovered, the patient is treated symptomatically. The Indian government has introduced the JE vaccination, which is administered to infants under the age of two. Children are given one dosage when they are 9 months old and another when they are 16 to 24 months old.

Dengue fever has symptoms like fever, headache, muscle and joint pain, rash, nausea, and vomiting. Some infections result in dengue hemorrhagic fever (DHF) or dengue shock syndrome (DSS). DSS has all the symptoms of DHF, along with patients having a rapid and weak pulse, narrow pulse pressure, and cold skin.

As of now, no specific antiviral drug or vaccine against dengue is available. The only solution is to control Aedes aegypti mosquitoes. The treatment is based on the signs and symptoms of the disease and confirmed after blood tests. Chikungunya is diagnosed with enzyme-linked immunosorbent assay (ELISA) blood tests. Because the symptoms of chikungunya and dengue fever are so similar, laboratory testing is crucial. Chikungunya does not have a specific treatment. Getting lots of rest and receiving supportive counseling for symptoms may be beneficial. This newly designed system stores data on various activities performed for the diagnosis and treatment of VBDs.

## 4.3 Framework view of (Diagnosis and treatment) VBDs

The work diagram given in figure 9, shows use case analysis of VBDs suffered cases, and treatment provided to them with the help of designed VBD ontology and SWRL rules.

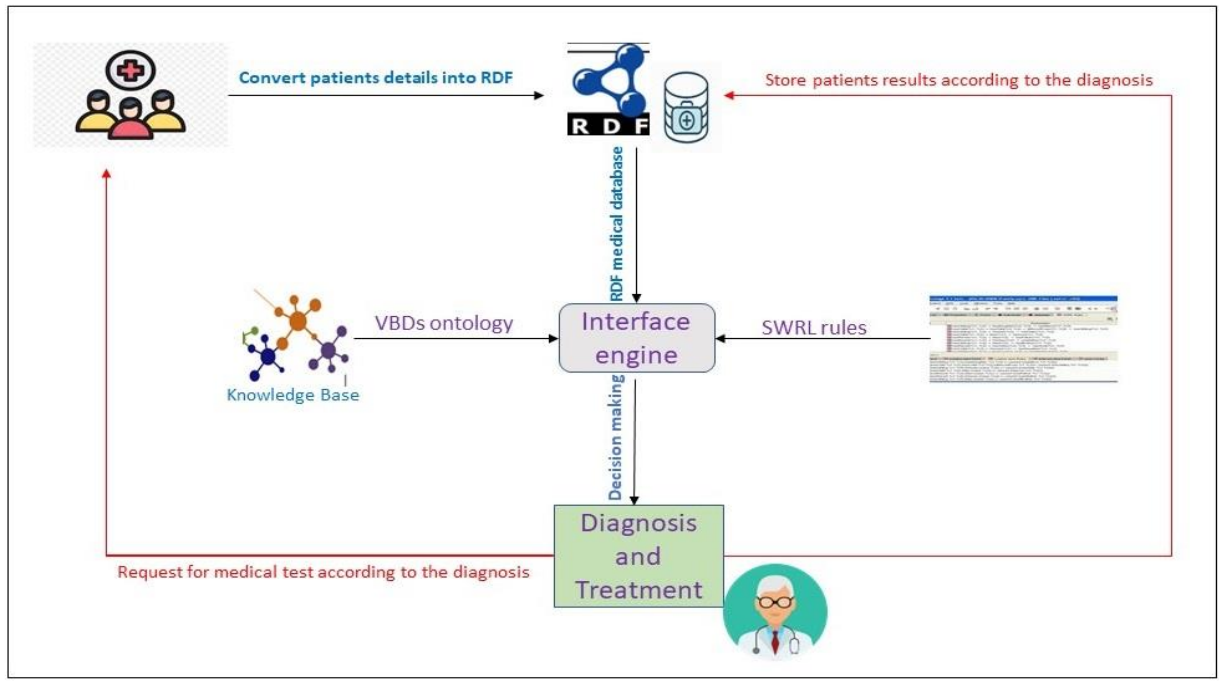

Figure 9: Use case analysis of VBDs patients based on VBDs ontology and SWRL rules.

The diagnosis and treatment provide suggestions to the patients for further precaution, health checkups (i.e., blood test, x-ray) and medicines based on diagnosed VBDs. First, the patient data is collected into RDF format and then uses VBDs ontology and SWRL rules on this data, then diagnosis and treatment is performed with the corresponding disease as shown in figure 19. During the complete process, Protege software [47] is working in the background. It is a java interface which helps in framework view of ontology results. Proteges have a reasoner inbuilt function which checks if the ontology was consistent or not. If it was incontinence then give warning, if not then give the inference output according to the SWRL rules. Many reasoners (i.e Jcel, Hermit etc.) are inbuilt in Protege, but in this work Pellet [46] is used. The complete working process shown in Figure 10.

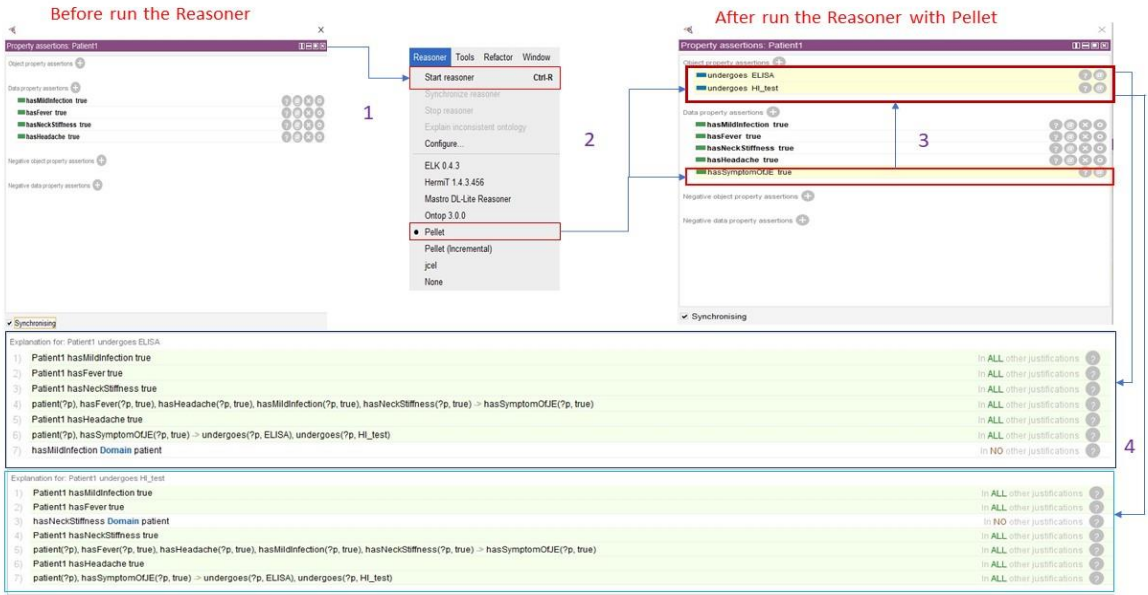

Figure 10: Framework view of Pellet working process, diagnosis, and treatment of patient1

In figure 10, step 1 shows the details of patient 1 before running the reagent. The results are in yellow according to the data that was asserted by patient 1 after running the pellet reasoner, according to the data that was fed in patient details given in step 2. The step 3 output depicts JE symptoms and recommends an ELISA and HI test, which are depicted in yellow. Step 4 includes the SWRL rules used for this output and a complete description of both tests, which were shown in two different boxes in the last of the figures.

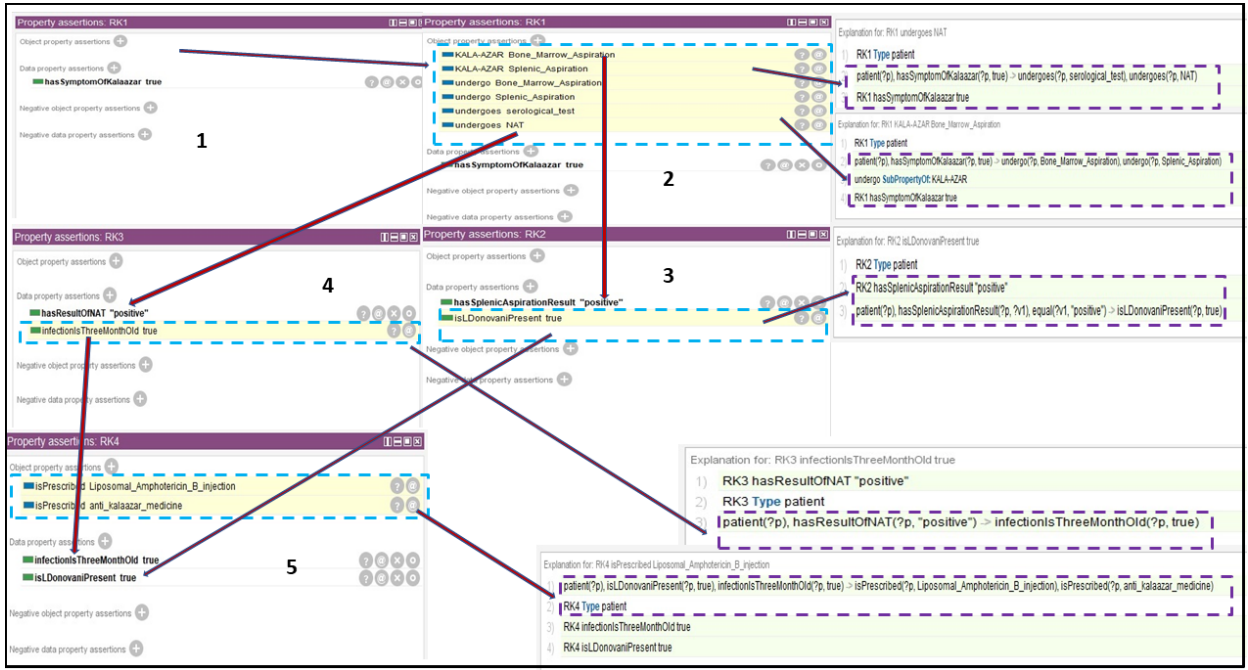

Figure 11: Process of diagnosis and treatment for example patient named as RK

Figure 11(1) shows a use case in which patient RK has Kala-Azar disease data and is now asserted to run the reasoner, and Figure 11(2) shows which tests require further analysis (shown in yellow) and which SWRL rules were used for those. RK has three tests prescribed: aspiration, NAT, and the serological (normal infection check) test [47]. In Figure 11(3), the result of the aspiration test is positive, which is asserted in the RK data. Again, run the reasoner and give the output in yellow, resulting in Leishmania donovani (L. donovani) being present in the body and also mentioning which SWRL rules will be used for this output. In Figure 11(4), the result of the NAT test is asserted to be positive, and then after running the reasoner, the output displays that there is a three-month-old infection and also suggests the SWRL rules responsible for this output. In Figure 11(5), the report asserts two tests, and the result after running the reasoner in yellow tells that liposomal amphotericin B injection and anti-Kala Azar medicine are prescribed. The respective SWRL rules are also presented in Figure 11. All other VBD diseases can also be diagnosed and treated in the same way.

## 5. Results and Discussion

The goal of this study is to collect biomedical text data on vector-borne diseases such as malaria, dengue, kala azar, and others from various sources such as doctor notes, medical mobile applications, and websites and convert this text data into RDF medical databases. Some of the VBDs' medical terms overlap due to the same medical checkup text terms or some other factors. To explain these overlapping terms, PCD ontology is used, which gives the proper relationship between all the terms. We collected thousands of text words across different VBDs and extracted meaningful words with the help of NLP, which is written in Table 6.

Table 6: Collection of text data for different VBDs

|  | Total text words collected after knowledge extraction model | Useful text words according to the PCD ontology | Accuracy % |
|---|---|---|---|
| Lymphatic Filariasis | 1987 | 1701 | 85% |
| Chikungunya | 2102 | 1986 | 94% |
| Dengue | 1806 | 1533 | 84% |
| Malaria | 2156 | 1987 | 92% |
| Kala-azar | 2400 | 2158 | 89% |
| Japanese Encephalitis | 1600 | 1561 | 97% |
|  | 12051 | 10923 |  |

A VBDs ontology is developed using BFO according to the NVBDCP guidelines, and to make this more accurate, an RDF medical database is added to the ontology, likewise adding different medical terms that are not available in the NVBDCP, and making it operational by adding individuals (i.e., patients) for diagnosis and treatment with the help of SWRL rules. A total of 72 SWRL rules are built for diagnosis and treatment. According to the RDF database, a total of 987 VBD patients have been diagnosed and treated, with 767 successfully treated and the remainder unsuccessful due to test reports that were not updated and text prediction that was unsuccessful due to missing text. The complete accuracy result information can be visualized in the form of a graph in figure 12.

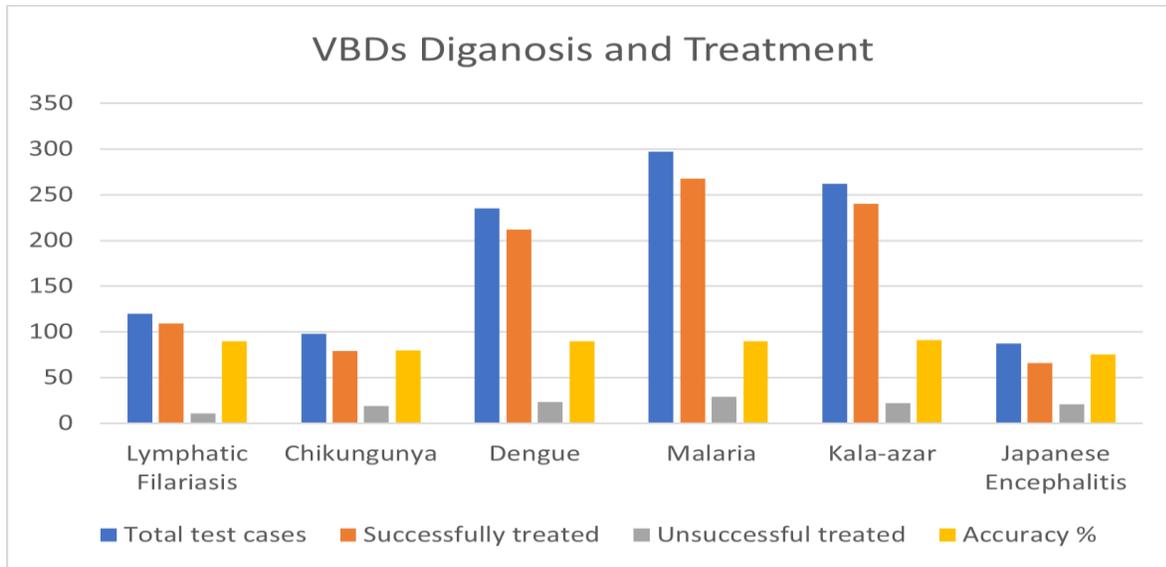

Figure 12: VBDs diagnosis and treatment results

Apart from this, our designed VBDs ontology gives evaluation based on available metrics count [48] as shown in Table 7.

Table 7: Total metrics count available in ontology

| Metrics | Value |
|---|---|
| Axiom | 6773 |
| Logical axiom count | 2604 |
| Declaration axioms count | 898 |
| Object property count | 153 |
| Data property count | 152 |
| Individual count | 987 |
| Annotation Property count | 25 |
| SubClassOf | 407 |
| DisjointClasses | 13 |
| SubObjectPropertyOf | 111 |
| ObjectPropertyDomain | 168 |
| ObjectPropertyRange | 291 |
| DataPropertyDomain | 138 |

*Schema metrics:* In terms of classes, attributes, relations, and individuals, the ontology could alternatively be described as a 5-tuple model.

O=<C, Dr, Sc, Re, Ind> where C-classes, Dr – data properties (attributes), Sc – subclasses, Ind – individuals, Re – Relations between classes.

Metrics can be evaluated based on the Attribute Richness, Relationship Richness, Class Richness and Average Population [49].

*Relationship richness:* Relationship Richness (RR) is a measure of the depth of connections between concepts in an ontology, and it is calculated with the help of equation 1.

$$RR = \frac{|Prop|}{|Subclass|+|Prop|} \quad \ldots\ldots\ldots\ldots (1)$$

Where |Prop| is the total number of properties, including attribute data and object characteristics (class relationships).

*Attribute richness:* As shown in equation 2, Attribute Richness (AR) is calculated by averaging the amount of attributes over the entire class.

$$AR = \frac{|Attribute|}{|Class|} \quad \ldots\ldots\ldots\ldots (2)$$

Where |attribute| is the total number of data attributes.

*Class richness (CR):* Class Richness (CR) is a sort of measurement that can be thought of as a knowledge metric because it indicates the whole amount of real-world knowledge conveyed through the created ontology. The CR is calculated with Equation 3 by dividing the number of classes with instances by the total number of classes.

$$CR = \frac{|Class\ with\_instance\ |}{|Class|} \quad \ldots\ldots\ldots (3)$$

*Average population (AP):* It figures out the average number of people in each class, which is shown in equation 4.

$$AP = \frac{|\ Individual\ |}{|\ Class\ |} \quad \ldots\ldots\ldots\ldots\ldots (4)$$

The schema metrics and their value with respect to VBDs ontology is represented in table 8.

Table 8: Different ontology metrics and their values

| Ontology Metrics | Value |
|---|---|
| RR | 0.6 |
| AR | 0.32 |
| CR | 0.46 |
| AP | 2.42 |

Also, check the ontology **quality score** through [49] based on the represented knowledge, which shows the domain knowledge's qualities and relationships according to equation 5.

$$\text{Score\_rk} = \frac{(|rel|*|class|*100) + (|sub-class|+|rel|)*|prop|}{(|sub-class|+|rel|)*|class|} \ldots\ldots\ldots\ldots (5)$$

Score can also be figured out based on how well base knowledge is taken out [48], as shown in equation 6.

$$\text{Score\_bk} = \frac{(Class\ with\_instance * 100) + (|Individual|)}{|class|} \ldots\ldots\ldots\ldots\ldots\ldots (6)$$

Ontology score is based on the formula in table 9 for score_rk and score_bk.

Table 9: Ontology Score Computation

| Evaluation parameter | Score |
|---|---|
| Score_rk | 86.05 |
| Score_bk | 92.03 |

Complete ontology view shown in figure 13.

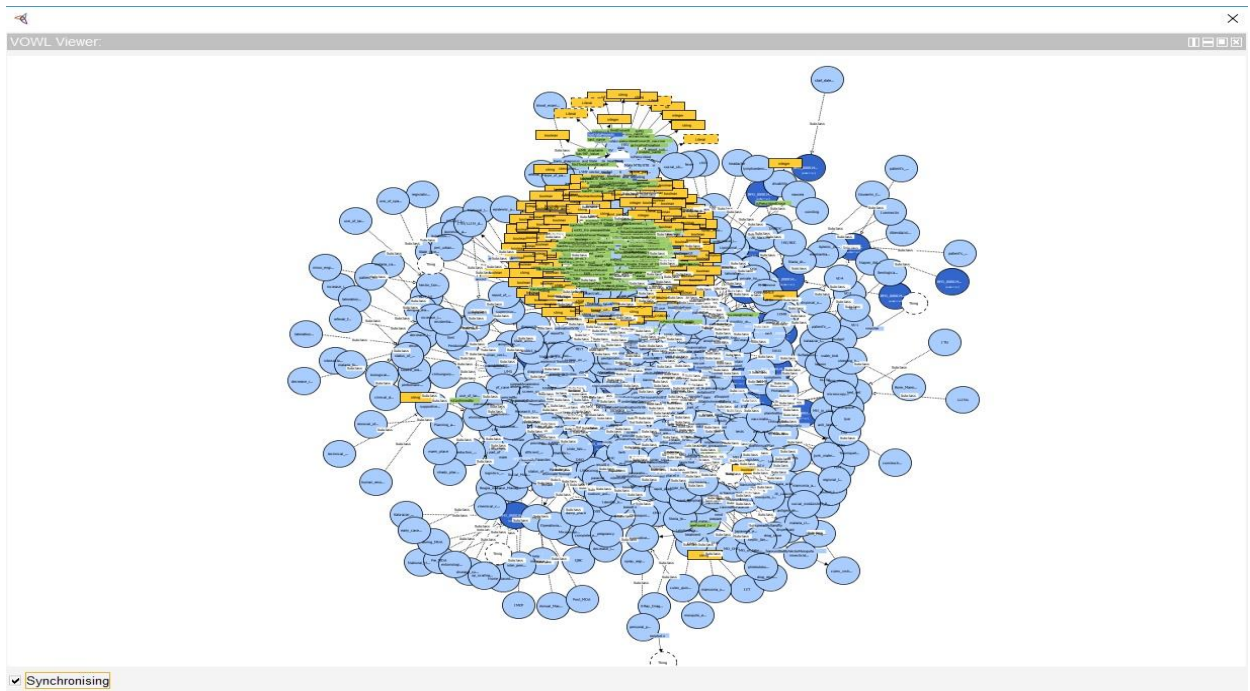

Figure 13: The complete ontology framework view shown VOWL in Protege 5.0. [50].

The diagnosis and treatment process suggested by the NVBDCP through its various guidelines has been implemented using SWRL. The diagnosis and treatment task, which lies in the hands of NVBDCP actors, is most likely to get valuable assistance from the proposed system. Also, queries on this VBDs ontology using DL and SPARQL [51] can be done for things like patient details, previous test reports of patients, precautionary guidelines, etc. It aids in decision-making based on the patient's health.

The time is also used to see how well SPARQL queries work. There were six queries we asked: Q1, Q2, Q3, Q4, Q5, and Q6. The query Q5 is hard because it has many parts. We run the same query on each of the four RDF medical data sets separately and look at the times in table 10 to see how well they work. Also, the RDF medical data that is used is put together and used for the same query. This is the best way to use time, since running the query on each set of data separately takes longer than running it on all of the data at once. Table 10 shows that it takes 2.1 seconds, 3 seconds, 3.9 seconds, and 4.8 seconds to process Q1 RDF medical data 1, 2, 3, and 4. Table 10 shows that it takes less time when all four processing times are added up and compared to the processing time for all RDF medical data. It found that our models and steps for estimating performance are the most likely to work. Aside from that, sometimes the query complexity is very high, but in those cases, our method gives the best performance, just like all other queries that are based on time. Even though it takes more time, putting all the data together gives us the best result, as shown in Figure. 14.

Table 10. Table showing the efficiency of medical-related queries on RDF data

| Query | RDF medical data 1 | RDF medical data 2 | RDF medical data 3 | RDF medical data 4 | Combine RDF medical data |
|---|---|---|---|---|---|
| Q1 | 2.1 | 3 | 3.9 | 4.8 | 5.8 |
| Q2 | 2.7 | 3.2 | 4.1 | 5.2 | 6.3 |
| Q3 | 2.4 | 3.3 | 4.3 | 5.4 | 6.2 |
| Q4 | 2.2 | 3.1 | 4.1 | 5.1 | 6.3 |
| Q5 | 5 | 6 | 7 | 8 | 9 |
| Q6 | 3 | 4.2 | 5.1 | 6.2 | 7.1 |

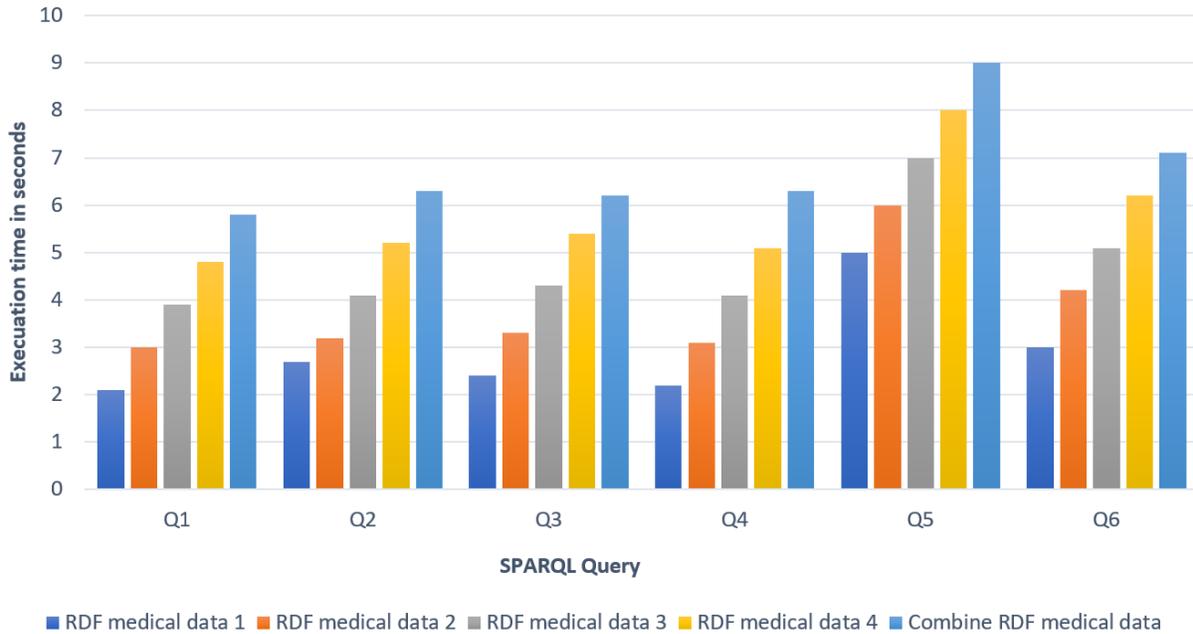

Figure 14. Evaluation of Graph-for-Query on Time-Varying RDF Medical Data

## 6. Conclusion

In this work, a VBD's ontology is developed, which handles all decision-making related to various VBDs. It also helps in the prevention and control of these kinds of diseases as per the NVBDCP guidelines. The created ontology can effectively represent the entire NVBDCP knowledge base, which contains the concepts behind the widely used NVBDCP program, using ontology graphs. The formalism used in this research will be very fruitful in the decision-making process, like case identification, diagnosis, prevention strategies, treatment, and recognizing the roles of various NVBDCP actors, etc. Also, through developed SWRL rules, it can improve the performance of the DSS. The OCR model is used here for text extraction from documents, and a spell checker is used for domain-specific meaning correction. The proposed DSS uses NLP for terminology extraction using NER, which is usually applied in PCD ontologies for converting text data into RDF. Our data extraction technology is making better use of patient information according to the PCD ontology. This can aid in medical diagnostics and enhance healthcare delivery by allowing for the inference and reasoning of new knowledge based on the VBD's ontology. It is also shown that the processing time of the system increases when dealing with large ontologies because of the massive amount of data, indicating that the scalability of the framework needs to be addressed. We are looking into adopting similar processing methodologies in this setting because they allow us to use many processors to examine different areas of the ontologies simultaneously, which in turn decreases execution time.

In the future, if there is any change in the guidelines or policies of the NVBDCP regarding the diagnosis and treatment of VBDs, it could also be implemented through our developed system by making some modifications. This work can be extended further for the cases of other vector-borne diseases by designing separate rules for diagnosis and testing, and these rules can be applied to the treatment regimen of the respective diseases. We can also extend the model's performance through the fusion of big data technologies with artificial intelligence.

## Acknowledgments

This research is supported by "Extra Mural Research (EMR) Government of India Fund by Council of Scientific & Industrial Research (CSIR)," Sanction letter no. – 60(0120)/19/EMR-II. The authors are grateful to CSIR for giving them the tools they needed to do the research. The authors are also grateful to the people in charge of the "Indian Institute of Information Technology, Allahabad at Prayagraj," which gave us the infrastructure and help we needed.